\pdfoutput=1

\documentclass[11pt]{article}

\usepackage[]{acl}

\usepackage{times}
\usepackage{latexsym}

\usepackage[T1]{fontenc}

\usepackage[utf8]{inputenc}
\usepackage{lettrine}
\usepackage{graphicx}

\usepackage{multirow}
\usepackage{array}
\usepackage{microtype}

\usepackage{enumitem} 
\setlist[itemize]{noitemsep, topsep=0pt}
\setlist[enumerate]{noitemsep, topsep=0pt}
\title{Quantifying Synthesis and Fusion 
and their \\Impact on Machine Translation}

\author{Arturo Oncevay${ }^{\epsilon}$\thanks{~~Work started when the first author was doing a research internship with JB at Aalborg University, Campus Copenhagen}~ \quad Duygu Ataman${ }^{\eta}$ \quad Niels van Berkel${ }^{\alpha,\mu}$ \quad Barry Haddow${ }^{\epsilon}$ \\ \textbf{Alexandra Birch${ }^{\epsilon}$ \quad Johannes Bjerva${ }^{\alpha,\sigma}$} \\
  ${ }^{\epsilon}$University of Edinburgh~~~${ }^{\eta}$New York University~~~
  ${ }^{\alpha}$Aalborg University (Aalborg$^{\mu}$ $|$ Copenhagen$^{\sigma}$) \\
  \texttt{a.oncevay@ed.ac.uk} \\}

\begin{document}
\maketitle
\begin{abstract}
Theoretical work in morphological typology offers the possibility of measuring morphological diversity on a continuous scale. 
However, literature in Natural Language Processing (NLP) typically labels a whole language with a strict type of morphology, e.g. fusional or agglutinative. 
In this work, we propose to reduce the rigidity of such claims, by quantifying morphological typology at the word and segment level. 
We consider \citet{payne-2017-morphological}'s approach to classify morphology using two indices: synthesis (e.g. analytic to polysynthetic) and fusion (agglutinative to fusional). For computing synthesis, we test unsupervised and supervised morphological segmentation methods for English, German and Turkish, whereas for fusion, we propose a semi-automatic method using Spanish as a case study. 
Then, we analyse the relationship between machine translation quality and the degree of synthesis and fusion at word (nouns and verbs for English-Turkish, and verbs in English-Spanish) and segment level (previous language pairs plus English-German in both directions). We complement the word-level analysis with human evaluation, and overall, we observe a consistent impact of both indexes on machine translation quality.
\end{abstract}

\section{Introduction}

One of the first barriers to develop language technologies is morphology, i.e., how systematically diverse their word formation processes are. For instance, agglutination and fusion are two morphological kind of processes that concatenate morphemes to a root with explicit or non-explicit boundaries, respectively. Processing morphologically-diverse languages and evaluating morphological competence in NLP models is relevant for language generation and understanding tasks, such as machine translation (MT). It is unfeasible to develop models with capacity large enough to encode the full vocabulary of every language, and it is a must to rely on subword segmentation approaches that help to constrain the capacity when generating rare, or even new words \cite{sennrich-etal-2016-neural}. Hence, understanding morphology is essential to develop robust subword-based models and evaluate the quality of their outputs \cite{vania-lopez-2017-characters}. Nevertheless, there is a potential gap between the probing of whether an NLP model can handle "morphological richness", and what is a proper measure of "morphological richness" from linguistic typology. 

In most of the recent NLP literature, different types of languages (e.g. agglutinative, polysynthetic) are chosen to test a more diverse handling of morphological richness \cite{ponti-etal-2019-modeling}.
There is, however, a debate as to whether languages can indeed be classified into discrete morphological categories. \citet{payne-2017-morphological} provided a morphological typology measurement in a  continuous spectrum using the indices of synthesis and fusion. 
Synthesis measures if a segment is highly analytic or synthetic (from 1 to more), whereas fusion measures whether it is highly agglutinative or fusional (from 0 to 1).
And surprisingly, with respect to NLP publications, it is possible to identify English segments with a very low fusion index, meaning that they are highly agglutinative\footnote{For instance, in the following fragment \cite{payne-2017-morphological}, the index of fusion is 1/8 or 0.125 (fusional morpheme joints are marked with a dot and the rest with a hyphen): "The company-'s great break-through came.PAST when they decid-ed to buy trike-s to sell their ice cream around the street-s in the nine-teen twenty-s".}. 

From a more applied perspective, if the references of an evaluation set (in any language generation task) are labelled with the indices, we could perform a stratified analysis (e.g. low fusion and high fusion) to determine how well an NLP model handles morphology for multiple languages. For example, we could assess whether a machine translation model is failing in generating more fusional than agglutinative segments for a specific target language. Knowing and quantifying that problem concerning morphology is the first step towards proposing a solution. 
Our contributions then are listed as follows:
\begin{itemize}
\itemsep 0em
    \item 
    We present the first computational quantification of synthesis and fusion using standard NLP evaluation sets. 
    \item We analyse the relationship between the two indices and machine translation quality at word-level, and observe that a higher degree of synthesis or fusion usually corresponds to less accurate translations in specific word types (studying nouns and verbs in English-Turkish, and verbs in English-Spanish). 
    \item We complement this evaluation with manual annotation of synthesis and fusion. 
    \item We extend the analysis at segment-level, using the aforementioned language pairs plus English-German in both directions, and identify that some synthesis and fusion-based predictors are significant for MT system outputs.

\end{itemize}
Furthermore, we release all the annotated data and evaluation results\footnote{ \url{https://github.com/aoncevay/quantifying-synthesis-and-fusion}}.

\section{Background and related work}

\subsection{Morphological typology}
The field of morphological typology characterises languages in terms of their word and sentence building strategies \cite{payne-2017-morphological}, such as agglutination or fusion. In current NLP literature, Turkish is labelled as a highly agglutinative language for the explicit boundaries between their morphemes, whereas Spanish is labelled as fusional for the opposite reason. 

However, early typological studies started to quantify these strategies with parameters, and avoided to characterise languages with a single type in a holistic way (e.g. \citet{sapir1921types,greenberg1960quantitative,comrie-1989-language}). In this context, \citet{payne-2017-morphological} recently highlighted the indices of synthesis and fusion, which are defined as follows. 

\subsubsection{Synthesis}

The index of synthesis offers a scale to contrast highly analytic or synthetic languages. This implies whether a word is composed by one (analytic) or several (synthetic) morphemes \cite{payne-2017-morphological}. Synthesis can be computed as the ratio of number of morphemes per words, it is closer to 1 when the language is more analytic (e.g. Mandarin, or English to a less degree), and gets higher the more synthetic the language is (e.g. Turkish, Inuktitut). Polysynthesis can be present when the synthesis degree is higher than 3, although the boundary is arguable. 
Besides, as we claim in this study, any language can present different levels of synthesis if we evaluate them at a more fine-grained level.

\subsubsection{Fusion}
Fusion is the ratio of the fusional morphemes joints\footnote{Or how many grammatical, syntactic and semantic features are joint. More than one feature can be fused in a single morpheme.} per the total number of joints. This index goes from 0 to 1, or from highly agglutinative (e.g. Turkish) to highly fusional (e.g. Spanish) cases. However, we noticed that the computation of fusion is complex to automatise. 
For instance, \citet{payne-2017-morphological} indicates potential cases to identify fusional joints, such as in prefixes, suffixes, infixes, circumfixes, compounding, non-concatenative processes (reduplication, apophony, substractive morphology) or autosegmental morphemes. Current automatic tools are not designed to identify these cases for most languages.  

\subsection{Morphological typology on NLP}

A survey by \citet{ponti-etal-2019-modeling}, on computational typology for NLP, pointed out that morphological knowledge is potentially helpful for analysing the difficulty in generation tasks such as language modelling and neural MT for both unsupervised and supervised settings. 
More specifically, they suggested that the degree of fusion (related to the index of fusion proposed by \citet{payne-2017-morphological}) impacts in the rate of less frequent words, which is a relevant parameter for generation tasks.

Besides, the studies that address morphological typology are related to either the development of morphological analysis systems or the evaluation of typologically diverse languages in terms of morphology (e.g. \citet{vania-lopez-2017-characters,xu-etal-2020-modeling}). However, the typology used to distinguish languages varies across different studies. For instance, \citet{vania-lopez-2017-characters} considers four phenomena to label languages: fusionality, agglutination, reduplication and root-pattern; whereas \citet{xu-etal-2020-modeling} considers more fine-grained elements such as affixation (prefixation, infixation and suffixation) or partial reduplication. Similarly, a fine-grained analysis on non-concatenative morphology for MT was performed by \citet{amrhein-sennrich-2021-suitable-subword}. It is important to note that none of the previous studies have addressed the phenomena as a continuous index but as discrete features. 

Furthermore, other studies refer only to morphological typological features as part of the task of typological feature prediction from linguistic databases \cite{bjerva-augenstein-2018-phonology,bjerva-etal-2019-probabilistic,bjerva-etal-2019-uncovering,bjerva-etal-2020-sigtyp,bjerva-augenstein-2021-typological}, and further applications of general typological concepts on MT are scarce and do not focus on morphology \cite{oncevay-etal-2020-bridging}.

\subsection{Morphological segmentation and analysis}

Morphological segmentation \cite{harris1951methods} aims to split a word into morphemes. There are both supervised (e.g. pointer generator networks  \cite{mager-etal-2020-tackling}) and unsupervised approaches (e.g. the Morfessor family of methods \cite{creutz-lagus-2002-unsupervised,poon-domingos-2009-unsupervised} or Adaptor Grammars  \cite{eskander-etal-2019-unsupervised}), where the former ones have outperformed the latter ones. 

Besides, the most widespread unsupervised segmentation methods (Byte-Pair-Encoding \cite[BPE;][]{sennrich-etal-2016-neural} and a method based on unigram language modelling \cite{kudo-2018-subword}) are not linked at all to morphological segmentation, but they are used to constrain the vocabulary size for neural generation tasks. 

Finally, it is important to note that the index of synthesis can be computed with a robust morphological analyser or segmentation model (to count the number of morphemes), but neither of them are built to compute the index of fusion directly.

\section{How to compute Synthesis and Fusion?}

\subsection{Synthesis: automatic computation}
\label{subsec:synthesis-comp}
To automatically compute the index of synthesis, we require to perform a robust morphological segmentation. A rule-based morphological analyser and disambiguator might be the best option if available (which we use later for Turkish in \S\ref{subsec:synthesis-word-analysis}), but for the purpose of the study, we compare well-known supervised and unsupervised methods:

\begin{itemize}
\itemsep 0em
    \item Byte-Pair-Encoding (BPE) and Unigram Language Model (uniLM)\footnote{We analysed several vocabulary sizes (4k, 8k, 16k, 32k, 64k) but report only the best one, which is 64k for all cases.}  
    from SentencePiece \cite{kudo-richardson-2018-sentencepiece}.
    \item Morfessor \cite{poon-domingos-2009-unsupervised}.
    \item Pointer Generator Network (PtrNet) from the implementation of \citet{mager-etal-2020-tackling}.
\end{itemize}

\subsubsection{Datasets and evaluation}

We used the CELEX dataset of segmented words for English and German \cite{steiner-2016-refurbishing,steiner-2017-merging}, where we randomly split training and evaluation data (80-10-10). Besides, for the unsupervised methods, we use the newscommentary-v15 \cite{barrault-etal-2019-findings} and EuroParl-v10 \cite{koehn-2005-europarl} corpora\footnote{Other languages like Danish are also available and was tested, but we did not report the results here as there are not complementary MT evaluation sets.}. 
Furthermore, we define two metrics to assess the performance on computing synthesis:
\begin{itemize}
\itemsep 0em
    \item Accuracy count: Evaluates if the number of obtained morphemes in the hypothesis segmentation is the same as in the reference.
    \item Exact segmentation precision: Analyses if the split morphemes are the same. We first perform an automatic alignment between the hypothesis and reference segments with the parallel Needleman-Wunsch algorithm for sequences \cite{naveed2005parallel}, and then compute the exact match at morpheme level.
   
\end{itemize}

\subsubsection{Results and discussion}

\begin{table}[t!]
\centering
\setlength{\tabcolsep}{2pt}
\resizebox{\linewidth}{!}{%
\begin{tabular}{l|cccc||cccc}
            & \multicolumn{4}{c||}{\textbf{English}}                                          & \multicolumn{4}{c}{\textbf{German}}                           \\ \hline
\#morphs.    & 1       & 2       & 3        & 4              & 1       & 2       & 3        & 4         \\ 
    & \small 16,914       & \small 28,900       & \small 1,798        & \small 73              & \small 13,061       & \small 32,007       & \small 5,808        & \small 360         \\ \hline \hline
            & \multicolumn{8}{c}{\textbf{Accuracy Count}}                                                                                                   \\
uniLM\textsubscript{64k} & 0.54          & 0.52          & 0.49          & 0.59                    & 0.35          & 0.27          & 0.21          & 0.18          \\
BPE\textsubscript{64k}   & 0.5           & 0.53          & 0.5           & 0.52                    & 0.29          & 0.33          & 0.28          & 0.26          \\
Morfessor   & 0.22          & 0.47          & 0.55          & 0.48                    & 0.17          & 0.26          & 0.28          & 0.25          \\
PtrNet  & \textbf{0.82} & \textbf{0.84} & \textbf{0.56} & \textbf{0.81}     & \textbf{0.74} & \textbf{0.86} & \textbf{0.7}  & \textbf{0.42} \\ \hline
            & \multicolumn{8}{c}{\textbf{Exact Segmentation Precision}}                                                                                                  \\
uniLM\textsubscript{64k} & 0.54          & 0.52          & 0.6           & 0.8             & 0.29          & 0.38          & 0.32          & 0.22          \\
BPE\textsubscript{64k}   & 0.5           & 0.44          & 0.56          & 0.76            & 0.24          & 0.33          & 0.23          & 0.08          \\
Morfessor   & 0.21          & 0.58          & \textbf{0.7}  & 0.78            & 0.17          & 0.45          & 0.44          & 0.36          \\
PtrNet    & \textbf{0.76} & 0.67          & \textbf{0.81} & \textbf{0.8}  & \textbf{0.67} & \textbf{0.73} & \textbf{0.72} & \textbf{0.62} \\ 

\end{tabular}
}
\caption{Accuracy count and segmentation precision for English and German using unsupervised and supervised segmentation methods. Results are grouped by the expected number of morphemes (e.g. "1" means that the word should not be split).}
\label{tab:index}
\end{table}

\begin{table*}[ht]
\setlength{\tabcolsep}{4pt}
\resizebox{\linewidth}{!}{
\begin{tabular}{p{2cm}|p{4cm}|p{3.9cm}|p{2.5cm}|p{1.85cm}|p{2.5cm}|p{1.5cm}|p{1cm}}
\multicolumn{8}{l}{Example (es): \textbf{Hablaremos} de la propuesta con la que se \textbf{condenó} a la ex primer ministra y fue \textbf{apoyada} por 147 diputados en la votación.}  \\ \hline
Verbs              & Features (spaCy) & Features (UniMorph)          & Segmentation          & feats. per morph          & fusional morph. joints          & total joints          & Fusion index     \\ \hline \hline
hablaremos \emph{(we will talk)}          & Mood=Ind, Number=Plur, Person=1, Tense=Fut, VerbForm=Fin &  V;IND;FUT;1;PL                 & habl - are - mos            & 0 -- 2\textsubscript{(IND;FUT)} -- 2\textsubscript{(1;PL)}                        & 0+(2-1)+(2-1) = 2                  & 2+2 = 4               & 0.5                  \\ \hline
condenó \emph{(condemned)}           &  Mood=Ind, Number=Sing, Person=3, Tense=Past, VerbForm=Fin & V;IND;PST;3;SG;PFV                 & conden - ó              & 0 - 5\textsubscript{(IND;PST;3;} \textsubscript{ SG;PFV)}                          & 0+(5-1) = 4                        & 4+1 = 5               & 0.8                   \\ \hline
apoyada \emph{(supported)}           &   Gender=Fem, Number=Sing, VerbForm=Part  & V.PTCP;PST;FEM;SG   & apoy - ada              & 0 - 3\textsubscript{(PST;FEM;SG)}                          & 0+(3-1) = 2                        & 2+1 = 3               & 0.66        \\ \hline          
\end{tabular}
}
\caption{Annotation example in Spanish. We first identify the verbs (in bold) and obtain their morphological features (using spaCy and the UniMorph schema). Then, we split the verb into its morphemes (segmentation), and identify which features are fused in each morpheme (feats. per morph). Finally, we compute the index of fusion by dividing the fusional morpheme joints by the total joints (which includes the agglutinative or explicit boundaries). On a side note, examples of verbs with zero fusion are in the infinitive (e.g. hablar \emph{(to talk)}) and gerund (e.g. hablando \emph{(talking)}) forms. }
\label{tab:annotation-example}
\end{table*}

Table \ref{tab:index} shows the scores on morphological segmentation for both English and German. We observe that both BPE and uniLM under-perform when it is not expected to split the word (column "1"). This is a pattern observed by \citet{bostrom-durrett-2020-byte}, where they noted that unsupervised segmentation methods tend to over-split the roots of words. They both improve their accuracy and precision when the number of expected morphemes is larger. Unexpectedly, Morfessor also under-performs in the "1" case for both languages, and only surpasses the other unsupervised methods when we measure precision for many morphemes. Furthermore, The PtrNet supervised method outperforms the rest in almost all scenarios. 

We conclude that, to compute synthesis, we should prioritise, besides a rule-based morphological analyser, a supervised segmentation method like PtrNet if data is available. We take advantage of this for the segment-level analysis in \S\ref{sec:segment-analysis}.

\subsection{Fusion: Semi-automatic computation}
\label{subsec:fusion-comp}

Calculating fusion should be approached in a case by case scenario, as there are different considerations provided by \citet{payne-2017-morphological}. Therefore, there is not an automatic tool that can obtain the fusion score directly. We decided to focus on Spanish\footnote{We chose this language because of the ease of finding annotators and MT training and evaluation data.} as a case study, where verbs and auxiliary verbs contains the highest degree of fusion of all the parts-of-speech (POS).

\paragraph{Procedure}

We observed that we could perform an annotation per paradigm and the termination of the verb (-ar, -er, -ir), as the fusion degree will remain the same regardless of the lemma\footnote{Except for irregular ones, which presents a limitation and potential noise. To reduce the risk of a biased assessment, we also performed a human evaluation.}. Then, on a chosen Spanish corpus:

\begin{enumerate}
\itemsep 0em

    \item Perform an automatic annotation of POS and morphological features\footnote{We use the spaCy model \texttt{es\_dep\_news\_trf}, available at \url{https://spacy.io/models/es\#es\_dep\_news\_trf}. It has an accuracy of 0.99 in POS and morphological tagging in the UD Spanish AnCora dataset \cite{taule-etal-2008-ancora}, which contains news texts mostly. }.
    \item Review the automatic annotation of special cases. For instance, there are 
    specific verb forms that are missed as adjectives. We corrected the POS and morphological annotation of those cases in a manual step.
    \item Obtain a set of all unique verb paradigms and morphological features in the corpus, considering the three different types of verb terminations in Spanish as different elements\footnote{Using the Unimorph database \cite{mccarthy-etal-2020-unimorph} is another alternative for extracting all the possible unique inflections. We aligned and considered both tag sets for the annotation, as shown in Table \ref{tab:annotation-example}.}.
\end{enumerate}

Now there is a list of unique verb paradigms and terminations 
that can be annotated both in synthesis and fusion. The steps are as follows:

\begin{enumerate}
    \itemsep 0em
    \item For each unique verb paradigm and termination, segment a verb sample into its morphemes. E.g. the verb \textit{habló} (`talked'), is split in \textit{habl-ó}, and \textit{habláramos} (`we were to speak') in \textit{habl-ára-mos}.
    \item Analyse how many morphological features are fused in each morpheme: if you change a value of a feature, will the surface form or morpheme will change? E.g. in \textit{habl-ó}, \textit{-ó} participates in 5  features (mode (indicative), subject person (third person), subject number (singular), tense (past) and aspect (perfective)). For \textit{habl-ára-mos}, \textit{-ára} includes the past and subjunctive, whereas \textit{-mos} denotes the person and number. If any of aforementioned feature changes its value, the surface will change too.
    \item Count and aggregate the results per morphemes and obtain the fusion for each verb paradigm. E.g. the fusion for \textit{habl-ó} is 4/5 = 0.8, and for \textit{habl-ára-mos} is 2/4 = 0.5.
\end{enumerate}

Finally, with the annotation in the unique list of verb inflections and terminations, we can extend the degree of fusion to all the verbs in the original Spanish corpus. An example of the annotation process is shown in Table \ref{tab:annotation-example}.

\begin{figure*}[t!]
    \centering
    \includegraphics[width=\textwidth]{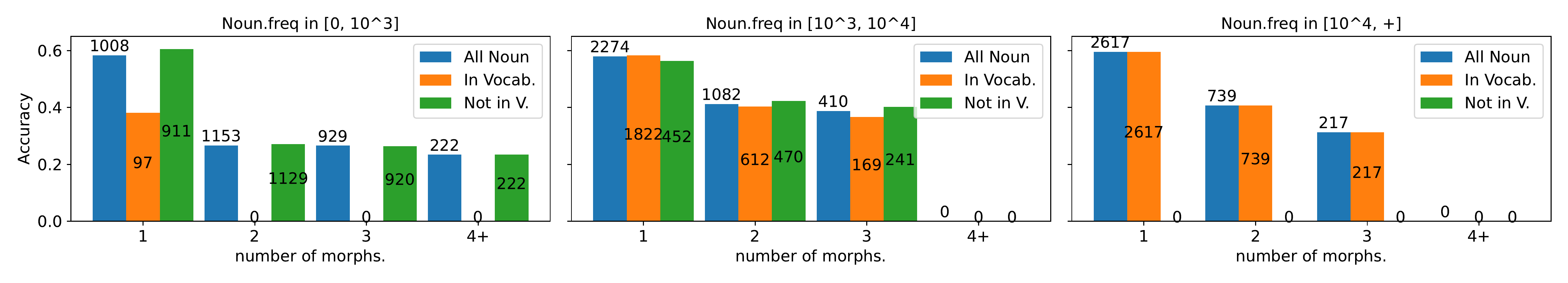}
    \includegraphics[width=\textwidth]{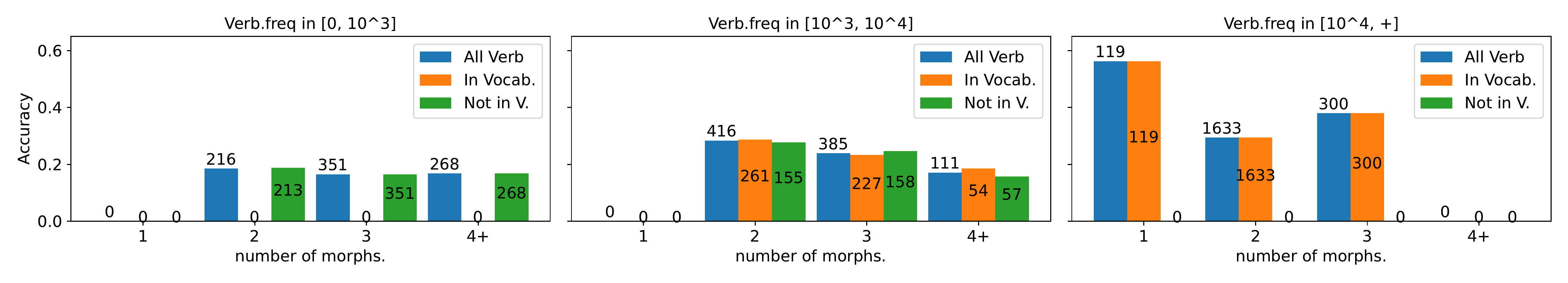}
    \caption{Accuracy (exact translation) for Nouns (top) and Verbs (bottom) in the English$\rightarrow$Turkish translations. Results are grouped by the training frequency of the words (less to more frequent from left to right), and each subplot presents the scores for all the words, and whether they belong or not to the vocabulary input of the model. The number of samples are stacked in each bar, and we do not show entries with less than 30 samples.}
    \label{fig:synthesis_turkish}
\end{figure*}

\section{Word-level analysis of Synthesis and Fusion in Machine Translation}

In this analysis, we ask the following question: how difficult is translating a word concerning its index of synthesis or fusion? For evaluating synthesis, we work with Turkish\footnote{Turkish presents high synthesis and agglutination \cite{Zingler+2018+415+447}, meaning that there are words composed with several morphemes and the morpheme boundaries are explicit, respectively. We focus on verbs and nouns, which usually contain more morphemes than other parts-of-speech. We chose this language due to the availability of an open-source rule-based morphological analyser and an expert annotator.} nouns and verbs, and for fusion, we keep working on Spanish verbs. For both cases, English is the source language in the translation task.

\subsection{Experimental design}
\label{sec:synthesis-exp-design}
The experiment consists of comparing a gold standard reference with machine translation system outputs at the word level:
\begin{enumerate}
\itemsep 0em
\item For both the reference and system output, we automatically \textbf{tag all the words with a morphological analyser} (the Boun morphological analyser and disambiguator \cite{sak2008turkish} for Turkish and an spaCy model trained on the Ancora Universal Dependency parser \cite{taule-etal-2008-ancora} for Spanish). The POS is needed to filter the target words. For synthesis in Turkish, the number of morphemes works as a proxy, as we are working at the word level. For fusion in Spanish, we need the inflection to obtain the degree of fusion from the annotated unique list (see \ref{subsec:fusion-comp}). 
\item \textbf{Align the words} between the reference and system output. We use the awesome-align \cite{dou-neubig-2021-word} tool by fine-tuning the multilingual BERT \cite{devlin-etal-2019-bert} model for word-alignment, using the reference and system output as parallel corpora.
\item Calculate the translation accuracy (exact match of the word, 0 or 1) for the target POS. 
\end{enumerate}
 We then fine-grain the results concerning the degree of synthesis (number of morphemes) or fusion. Additionally, we control different confounds: frequency of the word in the training set, and whether the full word is part of the vocabulary input of the model or not. Finally, we complement the analysis with a human evaluation (see \S\ref{subsec:human-eval}).

\subsection{Synthesis analysis: English$\rightarrow$Turkish}
\label{subsec:synthesis-word-analysis}

\paragraph{Data} We use the \textsc{newstest2018.en-tr} evaluation set from WMT \cite{bojar-etal-2018-findings}, with 3,000 samples. In the Turkish side there are 45,944 tokens, and Table \ref{tab:turkish-data} shows the distribution of the number of morphemes obtained with \citet{sak2008turkish}.

\paragraph{Model} We use an English-Turkish system trained with the TIL corpus of 39.9M parallel sentences \cite{mirzakhalov-etal-2021-large}.
On the \textsc{newstest2018.en-tr} set, the performance is 13.06 and 49.54 in BLEU and chrF, respectively.

\begin{table}[t!]
\centering
\setlength{\tabcolsep}{5pt}
\resizebox{\linewidth}{!}{%
\begin{tabular}{l|c|ccccc}
      & Total            & \#1    & \#2    & \#3    & \#4  & \#5+ \\ \hline
Verbs & 3,834    & 133   & 2,265 & 1,036 & 308 & 92 \\
Nouns & 10,680   & 5,899 & 2,974 & 1,556 & 244 & 7 
\end{tabular}
}
\caption{Number of nouns and verbs in the Turkish reference set, and their respective number of morphemes.}
\label{tab:turkish-data}
\end{table}

\paragraph{Results and discussion}
Figure \ref{fig:synthesis_turkish} shows the average accuracy (exact translation, 0 or 1) of nouns and verbs in \textsc{newstest2018.en-tr}, where the number of morphemes is a proxy for the index of synthesis. 
In most cases, especially with a higher training frequency, we observe that the average accuracy drops as the number of morphemes increases from 1 to more. This is clearer in nouns than in verbs, which have fewer cases to analyse overall. Between 2, 3 or more than 4 morphemes the differences are not significant, and sometimes is not consistent (e.g. verbs with the highest frequency). However, we can argue that analytic nouns (synthesis=1) are easier to translate than synthetic nouns (synthesis>1) for the English$\rightarrow$Turkish direction. The pattern holds for whether the word is part of the vocabulary of the model or not, although rare words (frequency in $[0, 10^3]$ have generally lower translation accuracy than more frequent words (frequency $ > 100$).

\subsection{Fusion analysis: English$\rightarrow$Spanish}
\label{subsec:fusion-word-spanish}

\paragraph{Data} We use the \textsc{newstest2013.en-es} evaluation set from WMT \cite{bojar-etal-2013-findings} with 3,000 samples. In the Spanish side there are 62,055 tokens, with 6,317 verbs, and where 1,411 of them are more agglutinative (fusion=0) and 4,822 more fusional (fusion>0). 

\paragraph{Model} For training, we use the MarianNMT toolkit \cite{junczys-dowmunt-etal-2018-marian-cost}, a Transfomer-base model  \cite{NIPS2017_7181} with default parameters, and four NVIDIA V100 GPUs. We obtained different English-Spanish models using the newscommentary-v8 \cite{bojar-etal-2013-findings} and EuroParl \cite{koehn-2005-europarl} datasets with joint vocabulary sizes of 8k, 16k and 32k (using unigram-LM from SentencePiece \cite{kudo-richardson-2018-sentencepiece}). For this analysis, we chose the best performing system: combining both datasets (2.2M sentences) with 16k pieces. On \textsc{newstest2013.en-es}, the performance is 31.6 BLEU points.

\begin{figure}[t!]
    \centering
    \includegraphics[width=\linewidth]{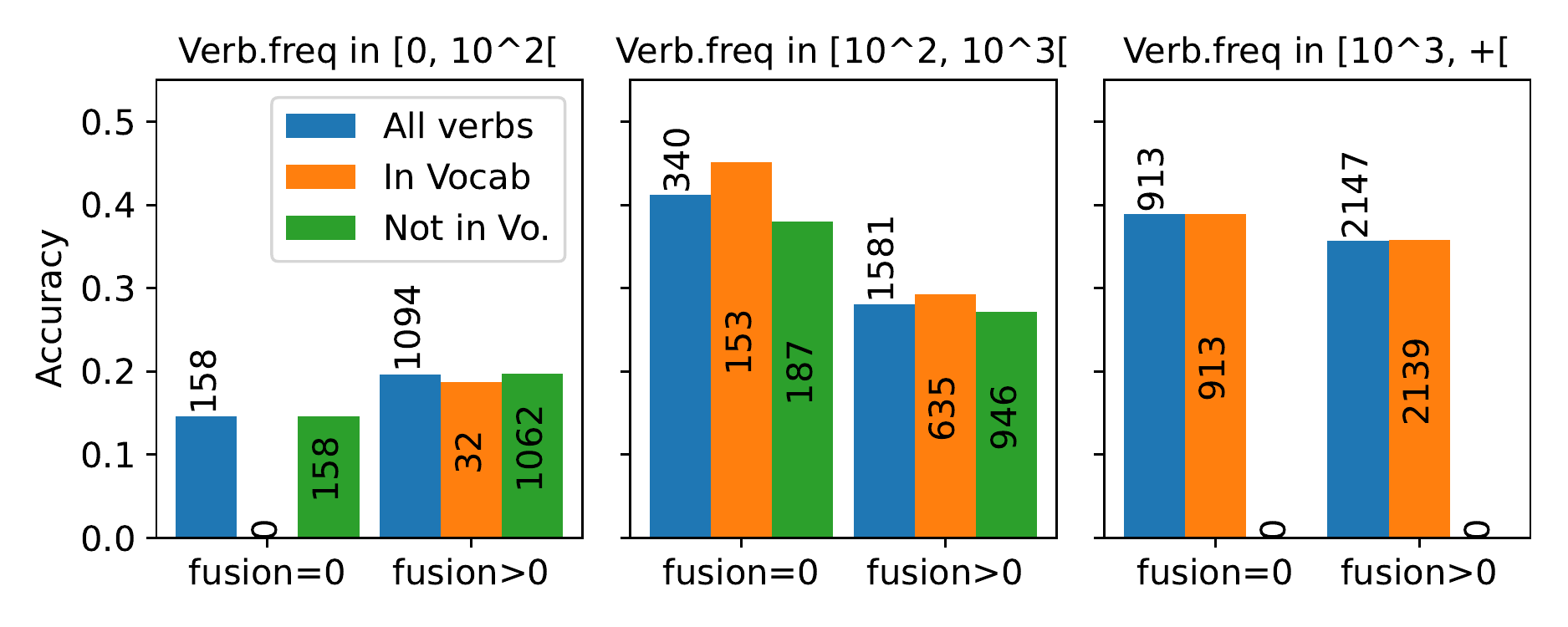}
    \caption{Accuracy (exact translation) for Verbs in the English$\rightarrow$Spanish translations. Results are grouped by the training frequency and whether the word belongs to the vocabulary of the model (In V) or not (Not in V).}
    \label{fig:fusion_spanish}
\end{figure}

\paragraph{Results and discussion}
Figure \ref{fig:fusion_spanish} shows the average accuracy of verbs in \textsc{newstest2013.en-es} for verbs without and with some degree of fusion. 
In the two higher frequency subplots (middle and right), we can observe that the average accuracy of the non-fusional verbs is higher than the fusional ones, and the pattern holds whether the verb is present in the vocabulary input of the model or not. The exception is for the least frequent verbs, although this is explained as the model do not have enough information to learn from, regardless of their degree of fusion.

\subsection{Human evaluation}
\label{subsec:human-eval}

Exact translation accuracy has limitations, as there are potential translations that could be acceptable given a specific context (e.g. a synonym). For that reason, we performed a human evaluation of a sample of sentences on (10\%) of each evaluation set, focusing on two scores\footnote{Details of the annotation protocol are in the Appendix}: 

\begin{enumerate}
\itemsep 0em
\item Semantic score: evaluates the meaning of the word used in the automatic translation (system output) and how it compares with the gold standard translation. Scale goes from 1 (no relationship at all) to 4 (it is the same lemma).
\item Grammar score: evaluates the grammatical form and how it compares with the gold standard translation. Scale goes from 1 (different inflection) to 3 (same inflection).
\end{enumerate}

\begin{figure*}[t!]
    \centering
    \includegraphics[trim={0.4cm 0 0.375cm 0},clip,width=0.49\linewidth]{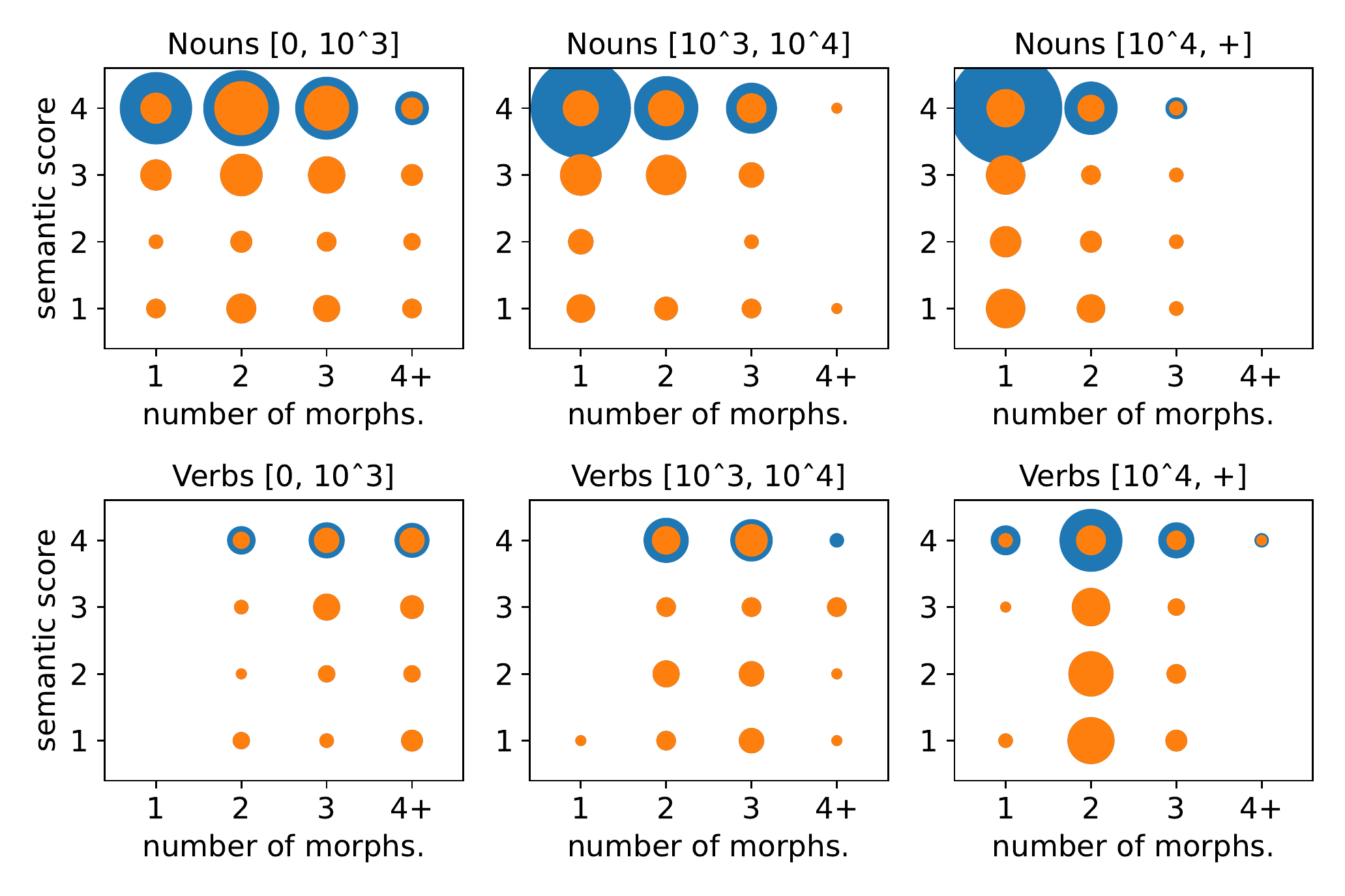}
    \includegraphics[trim={0.1cm 0 0.375cm 0},clip,width=0.49\linewidth]{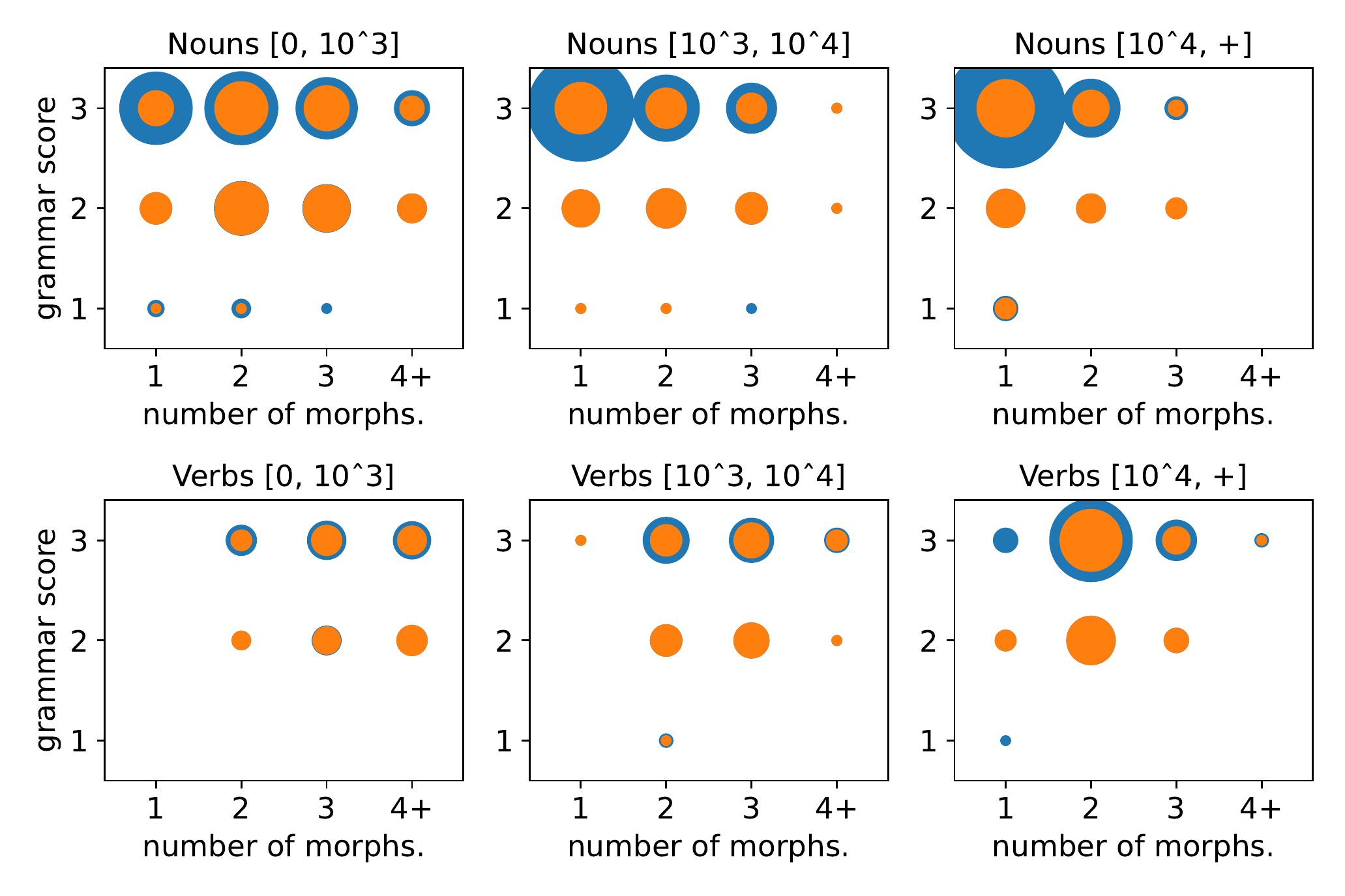}
\caption{Semantic score (left) and Grammar score (right) annotation for Turkish, for different frequency ranges of Nouns (top) and Verbs (bottom). Bubbles represent the amount of annotations per score and their respective group. The orange inner bubble represents the amount of samples with `zero' accuracy (in the automatic analysis) in each category.}
    \label{fig:tr_semantic_grammar}
\end{figure*}

\paragraph{Synthesis} In Figure \ref{fig:tr_semantic_grammar}, we show the annotation scores for the semantic and grammar metrics, for both nouns (top) and verbs (bottom). We also divide the analysis w.r.t. the frequency of the word in the training data. For nouns, we observe similar patterns as in the automatic analysis, where the amount of words with one morpheme (synthesis=1) has a higher semantic or grammar score than the rest, suggesting they are easier to generate for the model, except in the least frequent block, which still cannot be well translated. The verbs tend to have more distributed scores suggesting the difficulty of generating inflected forms may remain equally high even when the words are more frequent. Single morpheme verbs are very rare in Turkish and generally contain exceptional forms which reflects in the low translation accuracy (see Figure \ref{fig:synthesis_turkish}). 
We also observe that a good proportion of translated words with `zero' accuracy (not the exact translation, see the orange inner bubbles) has been annotated with highest semantic (same lemma) or grammar (same inflection) score, suggesting in some cases that the model is successful in generalization, although we see this case when the words are relatively short (1 to 3 morphemes).

\paragraph{Fusion}

\begin{figure}[t!]
    \centering
    \includegraphics[width=\columnwidth]{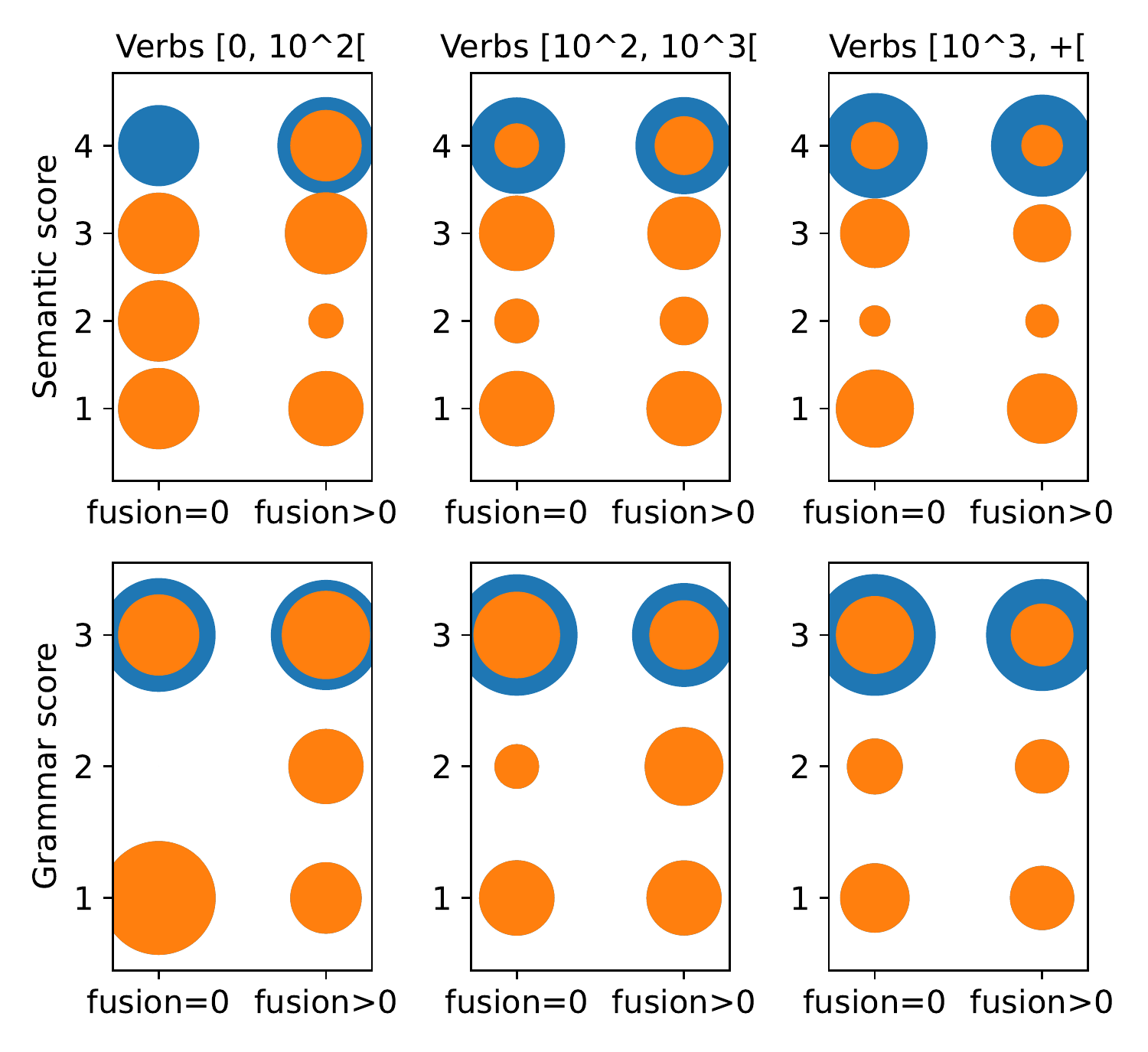}
    \caption{Semantic (top) and Grammar (bottom) annotation for Spanish.}
    \label{fig:es_semantic_grammar}
\end{figure}

Figure \ref{fig:es_semantic_grammar} shows the semantic and grammar annotation scores for Spanish verbs. For the semantic scores (top), in all levels, the gap between the non-fusional and fusional verbs is reduced, for all the frequency groups. This means that the model is indeed able to generalise and offer alternative translations (not the exact verb), which is more complex to measure with automatic metrics. In the grammar scale (bottom), however, we still note a slight advantage in the maximum score (3) of the non-fusional verbs against the fusional ones for the two highest frequency subplots (middle and right). This indicates that, with highly frequent verbs, it is still more difficult to translate correct forms with a fusion degree higher than zero. Similarly as for synthesis, we observe that there is a significant proportion of `zero' accuracy cases (orange inner bubbles) for the highest scores in most cases. This indicates that the model could generalise and translate verbs with similar meanings and not the exact, but close, forms.

\section{Segment-level Analysis of Synthesis and Fusion in Machine Translation}
\label{sec:segment-analysis}

Following up the word level analysis, we study the relationship between machine translation difficulty and the degree of synthesis or fusion at the segment level. For this purpose, we process a set of translation systems for the language pair we want to evaluate. The general steps are as follow:

\begin{enumerate}
\itemsep 0em
    \item For each system output, we compute \textbf{automatic evaluation metrics} (BLEU \cite{papineni-etal-2002-bleu}, chrF \cite{popovic-2015-chrf} and/or COMET \cite{rei-etal-2020-comet}) with respect to the reference set, per sentence.\footnote{Based on the analysis of \citet{kocmi-etal-2021-ship}, we prefer to report COMET and chrF over BLEU.}
    \item For each sentence of the evaluation set, we compute potential \textbf{predictor variables} for the automatic metric, such as the degree of synthesis or fusion. We complement the predictor variable list with other heuristics, such as the length of the sentence in characters (\textit{char.count}) or words (\textit{word.count}). The full list of all the predictors per language pair is in the Appendix.
    \item With the previous inputs, we generate \textbf{generalized linear models} per system output and evaluation metric, in which each model's output is set to the predictor variables. The goal is to identify which predictors affect each method's performance. 
    \item Following model creation, we extract the significant predictors of each model. This provides an indication of which variables can be used to predict the outcome of the model's dependent variable -- in our case the degree of synthesis or fusion, or any heuristic.\footnote{For simplification purposes, in the following analysis and plots, we only show the predictors that show  a significant effect on the system outputs.}
\end{enumerate}

\paragraph{Synthesis on En-Tr and Tr-En}

\begin{figure}[t!]
    \centering
    \includegraphics[trim={3cm 8.5cm 0 0},clip,width=\columnwidth]{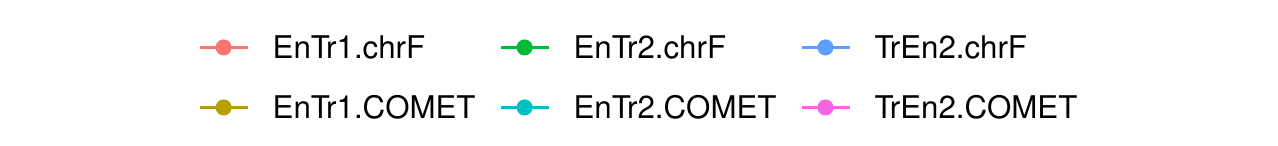}
    \includegraphics[trim={0.75cm 0 0 2cm},clip,width=\columnwidth]{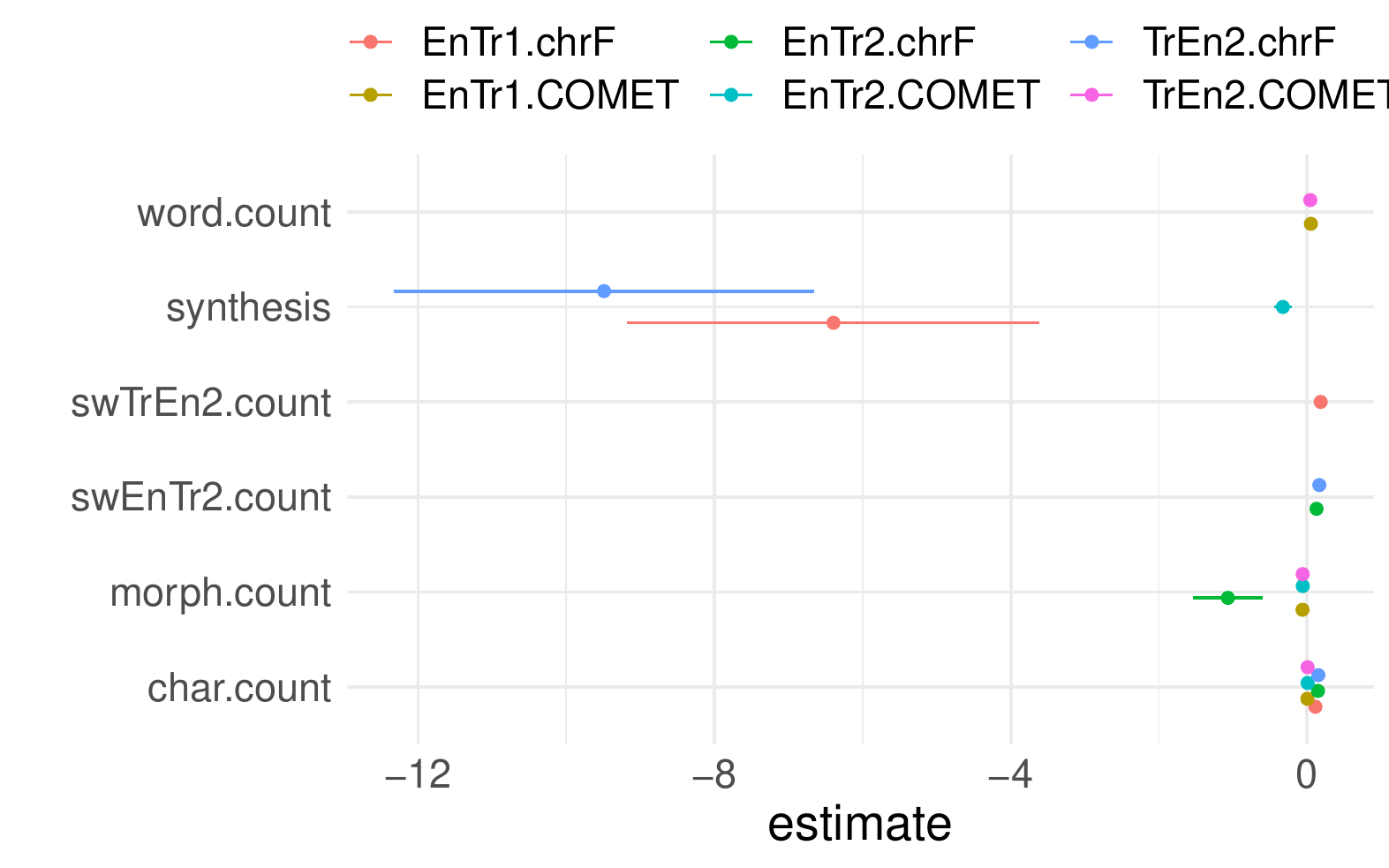}
    \caption{Overview of significant predictors for degree of synthesis across our TR-EN and EN-TR models.}
    \label{fig:sig_entr}
\end{figure}

We first start evaluating the English-Turkish and Turkish-English language pairs. The evaluated models are EnTr1, EnTr2, and TrEn2 (details in the Appendix). 
Also, as we are studying synthesis in Turkish, all predictors are computed on the Turkish side, regardless of the translation direction.

\begin{figure}[t!]
    \centering
    \includegraphics[trim={3cm 8.5cm 0 0},clip,width=\columnwidth]{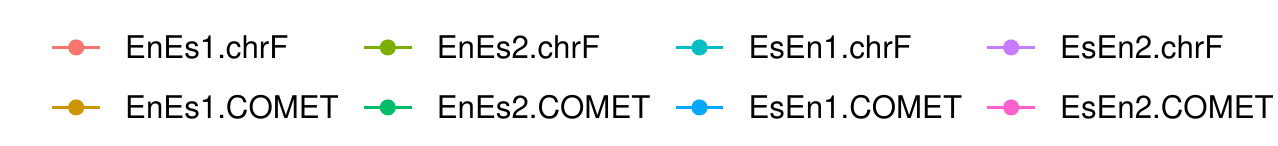}
    \includegraphics[trim={0.75cm 0 0.75cm 2cm},clip,width=\linewidth]{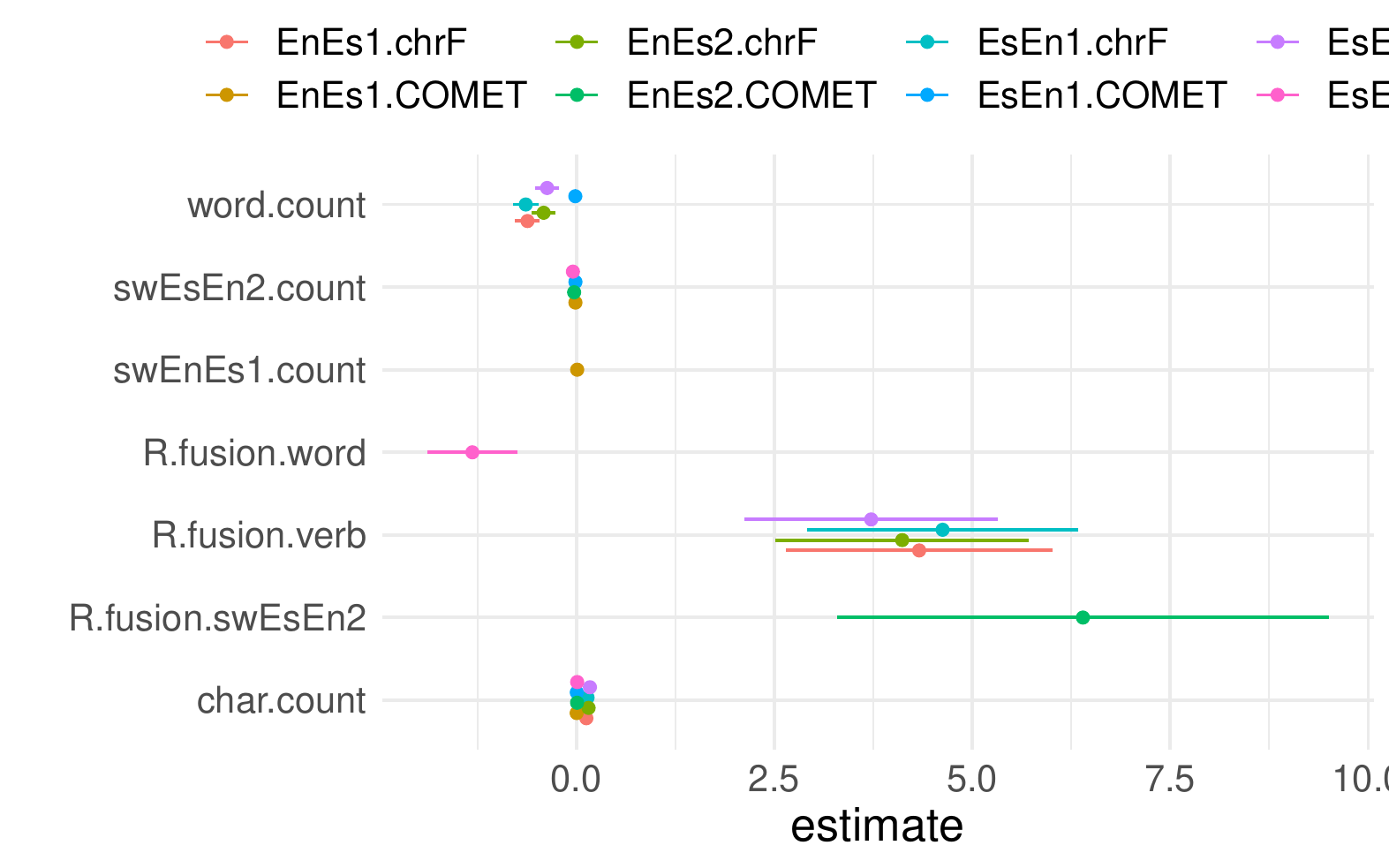}
    \caption{Overview of significant predictors for degree of fusion across our ES-EN and EN-ES models.}
    \label{fig:sig_enes}
\end{figure}

\begin{figure*}[t!]
    \centering
    \includegraphics[width=\linewidth]{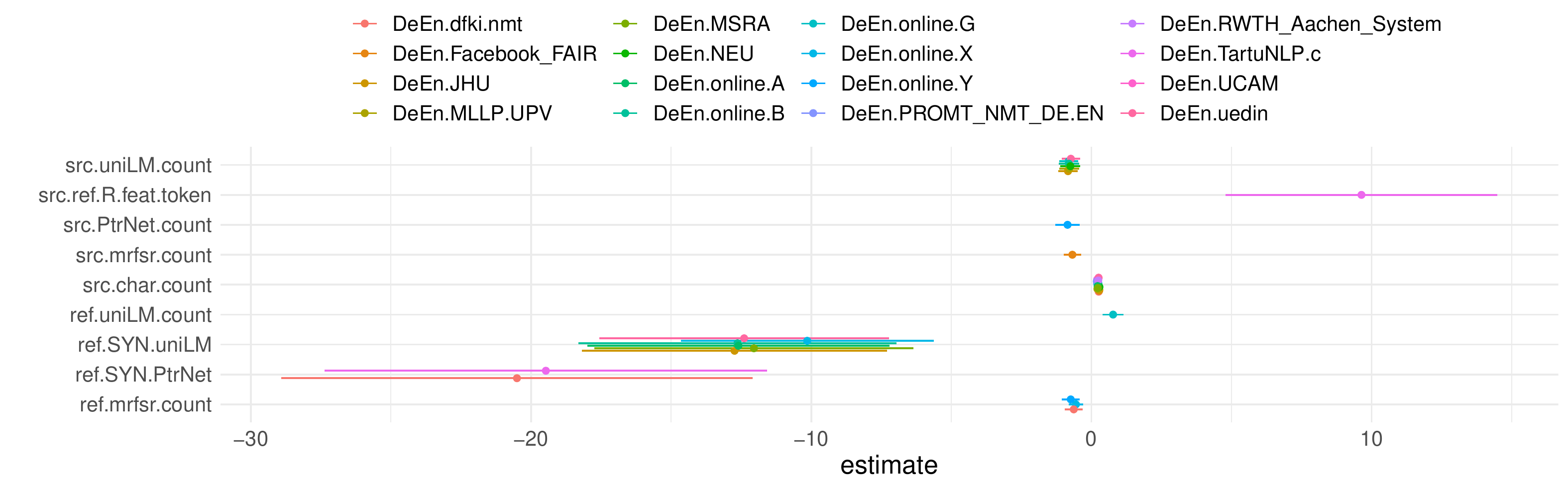}
    \caption{Overview of significant predictors across DE-EN models.}
    \label{fig:sig_deen}
\end{figure*}

Figure \ref{fig:sig_entr} presents an overview of the significant predictors on En-Tr and Tr-En systems, where we observe a large impact of the \textit{synthesis} variable on the chrF scores of two different systems (EnTr1 and TrEn2). The only other heuristic that achieves a notable impact on a system output is \textit{morph.count}, or the length of Turkish sentence in morphemes, split by a morphological analyser. Other predictors have a minor effect.

\paragraph{Fusion on En-Es and Es-En}

In a similar way, we evaluate the impact of fusion in English-Spanish (EnEs1, EnEs2) and Spanish-English (EsEn1, EsEn2) models (see the Appendix for details). 
Again, as we are studying fusion in Spanish, all predictors are computed on the Spanish side, regardless of the translation direction.

Figure \ref{fig:sig_enes} presents an overview of the significant predictors, where we observe that \textit{R.fusion.verb}, or the ratio of the degree of fusion over the number of verbs in the sentence, is the predictor that has the highest impact in most system outputs (EnEs1, EnEs2 and EsEn2). Additionally, \textit{R.fusion.swEsEn2} (or the ratio of the degree of fusion over the number of subwords input in the EsEn2 model) also has a high impact in one system output (EnEs2, which uses the same segmentation model as EsEn2).

\paragraph{Analysis on En-De and De-En}
Finally, we extend the analysis to English-German and German-English language pairs, using the respective evaluation sets of the WMT2018 campaign \cite{bojar-etal-2018-findings}, and the system outputs provided for all the participants (measured in BLEU). For computing synthesis, we use the different segmentation methods we compared in \S\ref{subsec:synthesis-comp}. However, for fusion, we only use a shallow proxy with the number of morphological features that are tagged using a morphological analyser. In this case, the predictors are computed for both the source and target side. 

We present an overview of these significant predictors for German-English in Figure~\ref{fig:sig_deen} (and the Appendix contains the results for English-German in Figure~\ref{fig:sig_ende}). We can observe that \textit{ref.SYN.uniLM} and \textit{ref.SYN.PtrNet} are the predictors that impact most of the different system outputs. These variables refer to the synthesis computed on the reference side (English) using uniLM or PtrNet as the morpheme segmentation method, respectively. Furthermore, we observe that \textit{src-ref.R.feat.token} has also some effect over one system output,
which is a shallow proxy for the fusion degree in the source w.r.t. to reference segment (using the ratio of number of features per number of tokens).

\section{Discussion}

It is important to note the limitations of this study. Overall results do not suggest that translating into more analytic languages (e.g. Chinese) or more agglutinative ones (e.g. Turkish) is easier than their counterparts. Highly analytic ones present the significant issue of word coverage and vocabulary size of the model. Besides, 
we cannot isolate the fusional degree from synthesis entirely. For instance, Turkish is a highly agglutinative language, but also highly synthetic, and there are languages that present both agglutinative and fusional traits, like Navajo. Moreover, the language scope is another limitation: is it possible to extend it to further languages in a practical way? Synthesis can be calculated directly only if the morphological analyser splits the word into morphemes, and fusion poses several issues as mentioned before. Furthermore,  \citet{payne-2017-morphological} also indicated that the discourse can impact the computed degrees due to the diversity of the vocabulary. This study focuses on news data only, and it will be relevant to extend it to a multi-domain approach.

To address the limitations, we consider that our word level analysis, that targets specific POS, has been fundamental to allow the study of the indexes, and to partially isolate them from each other (e.g. Spanish verbs do not present more than three morphemes, keeping a low synthesis value across all the analysis). Moreover, to rapidly extend the evaluation for new languages and domains, we could follow a less fine-grained analysis in each index. For instance, we can compare synthesis=1 vs. synthesis>1, or fusion=0 vs. fusion>0, as in this work.

\section{Conclusion and future work}
In conclusion, we proposed methods to quantify the indices of synthesis and fusion in automatic and semi-automatic ways, respectively. Besides, for the chosen language pairs, we observed that the studied degrees have an impact in machine translation performance at both word and segment level, where we included a human evaluation of the former case. 

Our analysis opens the possibility for new fine-grain evaluation approaches for MT and other NLP generation tasks. For instance, as future work, we can ask: are we improving the automatic translation of highly fusional words? or, are our proposed models more aware of fusional joints (non explicit boundaries)? Following our methodology, that targets specific POS tags, could aid to analyse whether new models are improving their performance in highly fusional words or segments. This could also be helpful for evaluation approaches in morphological segmentation. Furthermore, another potential research avenue is to aid model training for MT: e.g. knowing which segments are more or less synthetic and/or fusional could be beneficial for sampling strategies.

\section{Ethical Considerations}

The annotations in this paper were compensated accordingly (see Appendix). Also, for all the datasets used in the research, we stick to the ethical standards giving credit to the original author. We encourage future work that take advantage of these resources, to cite also the original sources of the data. We also see other ethical risks of this work: for the down-stream task of MT, a translation system should not be deployed with low quality translations, as it can mislead the user, and have implicit biases. 

\section*{Acknowledgements}

\lettrine[image=true, lines=2, findent=1ex, nindent=0ex, loversize=.15]{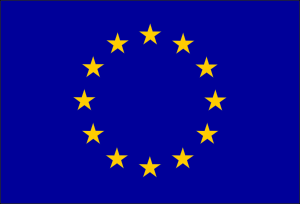}%
{T}his work was supported by funding from the European Union's Horizon 2020 research and innovation programme under grant agreements No 825299 (GoURMET) and the EPSRC fellowship grant EP/S001271/1 (MTStretch). Also, it was performed using resources provided by the Cambridge Service for Data Driven Discovery (CSD3) operated by the University of Cambridge Research Computing Service (\url{http://www.csd3.cam.ac.uk/}), provided by Dell EMC and Intel using Tier-2 funding from the Engineering and Physical Sciences Research Council (capital grant EP/P020259/1), and DiRAC funding from the Science and Technology Facilities Council (\url{www.dirac.ac.uk}). We express our thanks to Kenneth Heafield and Rico Sennrich, who provided us with access to the computing resources. 

Moreover, the first author was granted financial support from the European Association for Machine Translation (EAMT), under its programme “2020 Sponsorship of Activities”, and from the European Cooperation in Science and Technology (COST) under the programme CA18231 - Multi3Generation: Multi-task, Multilingual, Multi-modal Language Generation.


\bibliography{custom,typ4nlp}
\bibliographystyle{acl_natbib}

\appendix

\section{Human Evaluation}

\subsection{Annotation Protocol}
This study measures the translation quality of translations generated by a translation system. You are given a list of sentences where one column lists each word in the gold standard (correct) translation and the corresponding column the system-generated translations. The evaluation of the translations will rely on the two scores described below. The scores to use in the evaluation are:

\paragraph{Semantic score} evaluates the meaning of the word used in the automatic translation (system output) and how it compares with the gold standard translation.

Please assign each word in the output one of the scores you find most appropriate:
\begin{enumerate}
    \item There is no relationship between the two lemmas
    \item The lemmas are different but the translation does not fit well in the context
    \item The lemmas are different but it is still an acceptable translation (e.g. synonym)
    \item It is the same lemma
\end{enumerate}

\paragraph{Grammar score} evaluates the grammatical form and how it compares with the gold standard translation.

Please assign each word in the output one of the scores you find most appropriate:
\begin{enumerate}
    \item The word is inflected in a different way and it is not necessarily correct
    \item The word has different inflection but it is still grammatically correct
    \item The words have the same inflection, and it is correct
\end{enumerate}

Please annotate all words in the translations in the file shared with you. In your evaluation try assigning the two scores to each word independently. The inflection of the word measures the morphological feature and should also be evaluated independently from the analyzer output which is automated and may contain errors.

The file contains example annotations for your reference, please ask any questions related to unresolved annotation examples by contacting the project coordinators.

\subsection{Annotators}

For both Turkish and Spanish, the annotators were contacted directly due to their expertise in morphology (both of them are PhD students in Linguistics and Computational Linguistics, respectively), besides requiring that they are native speakers of the target languages. Also, they were paid more than the minimum wage per hour of annotation of their country of residence, and were told that the annotated data will be released upon acceptance of the study.

\section{Segment-level Analysis of Synthesis and Fusion }
\subsection{List of machine translation systems}

\begin{itemize}
    \item EnTr1: the same system used in \S4.2
    \item EnTr2: Transformer-base model \cite{NIPS2017_7181} with joint vocabulary size of 8k pieces (unigram language modelling from SentencePiece \cite{kudo-richardson-2018-sentencepiece}, and trained with a sample (10\%) of the corpus of EnTr1.
    \item EnEs1: the same system used in \S4.3
    \item EsEn1: similar configuration than EnEs1 but in the opposite direction
    \item EnEs2: same configuration as EnEs1 (model and vocabulary) but with smaller training data. It uses only newscommentary-v8 data, with around 300k sentences).
    \item EsEn2: similar configuration than EnEs2 but in the opposite direction.
\end{itemize}

\subsection{List of predictors}

Tables \ref{tab:pred-entr-tren}, \ref{tab:pred-enes-esen} and \ref{tab:pred-ende-deen} describes all the predictors used at the segment level analysis of English-Turkish, English-Spanish and English-German (both directions), respectively.

\begin{table}[]
\setlength{\tabcolsep}{2pt}
\resizebox{\linewidth}{!}{%
\begin{tabular}{l|l}
Predictor & Description \\ \hline
char.count & number of characters \\
word.count & number of words (no punct. or numbers) \\
morph.count & number of morphemes. \\
synthesis & ratio of morph.count / word.count \\
N+V.word.count & number of Nouns and Verbs \\
N+V.morph.count & number of morphemes of the Nouns and Verbs \\
N+V.synthesis & ratio of N+V.morph.count / word.count \\
swEnTr1.count & number of subwords processed by the EnTr1 model \\
swEnTr2.count & number of subwords processed by the EnTr2 model \\
swTrEn2.count & number of subwords processed by the TrEn2 model \\
syn.swEnTr1 & ratio of swEnTr1.count / word.count (synthesis proxy) \\
syn.swEnTr2 & ratio of swEnTr1.count / word.count (synthesis proxy) \\
syn.swTrEn2 & ratio of swEnTr1.count / word.count (synthesis proxy) \\
\end{tabular}
}
\caption{List of predictors for En-Tr and Tr-En. All variables are computed on the Turkish segment of the evaluation set.}
\label{tab:pred-entr-tren}
\end{table}

\begin{table}[]
\setlength{\tabcolsep}{2pt}
\resizebox{\linewidth}{!}{%
\begin{tabular}{l|l}
Predictor & Description \\ \hline
char.count & number of characters \\
word.count & number of words (no punct. or numbers) \\
verb.count & number of verbs \\
fusion & sum of the degree of fusion of all the verbs in the segment \\
R.fusion.verb & ratio of fusion / verb.count \\
R.fusion.word & ratio of fusion / word.count \\
swEsEn1.count & number of subwords processed by the EsEn1 model \\
swEsEn2.count & number of subwords processed by the EsEn2 model \\
R.fusion.swEsEn1 & ratio of fusion / swEsEn1.count \\
R.fusion.swEsEn2 & ratio of fusion / swEsEn2.count \\
swEnEs1.count & number of subwords processed by the EnEs1 model \\
swEnEs2.count & number of subwords processed by the EnEs2 model \\
R.fusion.swEnEs1 & ratio of fusion / swEnEs1.count \\
R.fusion.swEnEs2 & ratio of fusion / swEnEs2.count \\
\end{tabular}
}
\caption{List of predictors for En-Es and Es-En. All variables are computed on the Spanish segment of the evaluation set.}
\label{tab:pred-enes-esen}
\end{table}

\begin{table}[]
\setlength{\tabcolsep}{2pt}
\resizebox{\linewidth}{!}{%
\begin{tabular}{l|l}
Predictor & Description \\ \hline
src.char.count & number of characters in the source side \\
ref.char.count & number of characters in the target side \\
src.word.count & number of words in the source side \\
ref.word.count & number of words in the target side \\
src.uniLM.count & number of subwords obtained by uniLM in the source \\
ref.uniLM.count & number of subwords obtained by uniLM in the target \\
src.SYN.uniLM & synthesis in source = src.uniLM.count / src.word.count \\
ref.SYN.uniLM & synthesis in target = ref.uniLM.count / ref.word.count \\
src.mrfsr.count & number of subwords obtained by Morfessor in the source \\
ref.mrfsr.count & number of subwords obtained by Morfessor in the target \\
src.SYN.mrfsr & synthesis in source = src.mrfsr.count / src.word.count \\
ref.SYN.mrfsr & synthesis in target = ref.mrfsr.count / ref.word.count \\
src.PtrNet.count & number of subwords obtained by PtrNet in the source \\
ref.PtrNet.count & number of subwords obtained by PtrNet in the target \\
src.SYN.PtrNet & synthesis in source = src.PtrNet.count / src.word.count \\
ref.SYN.PtrNet & synthesis in target = ref.PtrNet.count / ref.word.count \\
src.feat.count & number of morph. features in the source (using spAcy) \\
src.R.feat.token & ratio of src.feat.count / src.word.count \\
ref.feat.count & number of morph. features in the target (using spAcy) \\
ref.R.feat.token & ratio of ref.feat.count / ref.word.count \\
src-ref.feat.count & src.feat.count minus ref.feat.count \\
src-ref.R.feat.token & src.R.feat.token minus ref.R.feat.token \\
ref-src.feat.count & ref.feat.count minus src.feat.count \\
ref-src.R.feat.token & ref.R.feat.token minus src.R.feat.token \\
\end{tabular}
}
\caption{List of predictors for En-De and De-En. Variables are computed either on source (src) or target (ref) side.}
\label{tab:pred-ende-deen}
\end{table}

\subsection{Results on English-German}

\begin{figure*}[h!]
    \centering
    \includegraphics[width=\linewidth]{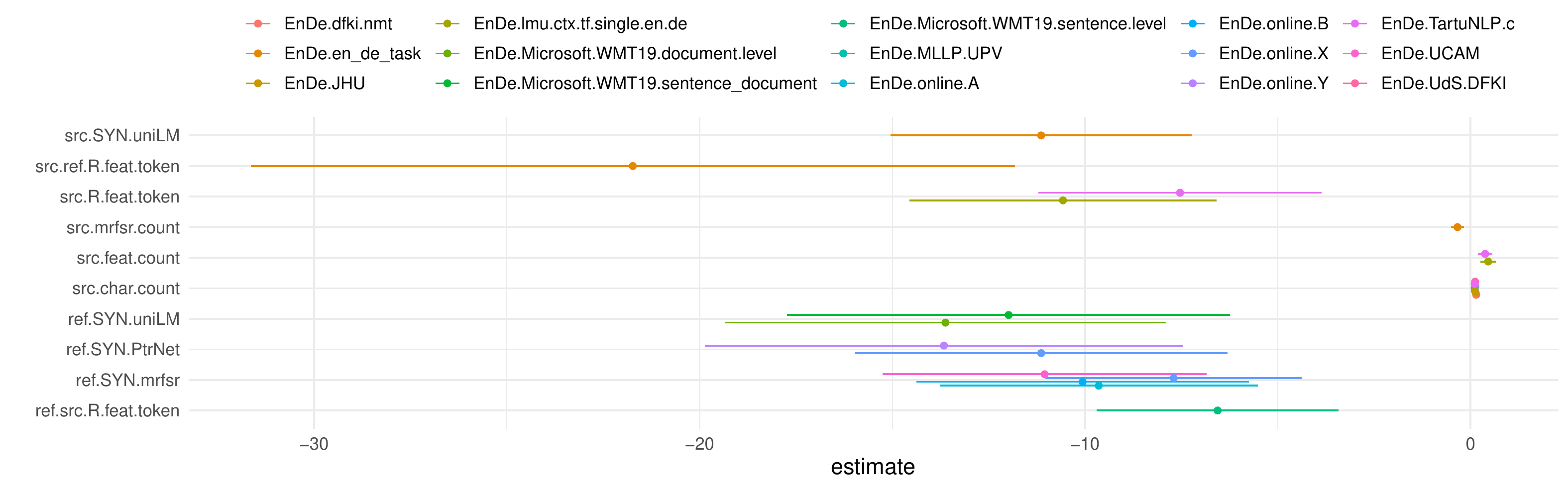}
    \caption{Overview of significant predictors for degree of synthesis across EN-DE models.}
    \label{fig:sig_ende}
\end{figure*}

Figure \ref{fig:sig_ende} shows the analogous results for English to German, where the synthesis-based variables presents a high impact w.r.t. the other predictors.



\end{document}